\documentclass[letterpaper]{article} 
\usepackage[preprint]{aaai2027}  
\usepackage[hyphens]{url}  
\usepackage{graphicx} 
\usepackage{natbib}  
\usepackage{caption} 
\usepackage{algorithm}
\usepackage{algorithmic}

\usepackage{newfloat}
\usepackage{listings}
\DeclareCaptionStyle{ruled}{labelfont=normalfont,labelsep=colon,strut=off} 
\floatstyle{ruled}
\newfloat{listing}{tb}{lst}{}
\floatname{listing}{Listing}

\usepackage{booktabs}

\usepackage{amsmath}
\usepackage{amssymb}
\usepackage{array}
\usepackage{colortbl}
\usepackage{makecell}
\usepackage{multirow}

\title{CF-LoRA: Decoupled Factor Aggregation and Adaptation-Aware Client Clustering for Federated LoRA Fine-Tuning}
\author{
    Mengjun Yi\textsuperscript{\rm 1,\rm 2},
    Langxing Yang\textsuperscript{\rm 1,\rm 2},
    Suhan Guo\textsuperscript{\rm 1,\rm 2},
    Furao Shen\textsuperscript{\rm 1,\rm 2}\corresponding,
    Jian Zhao\textsuperscript{\rm 3}
}
\affiliations{
    \textsuperscript{\rm 1}State Key Laboratory for Novel Software Technology\\
    \textsuperscript{\rm 2}School of Artificial Intelligence\\
    \textsuperscript{\rm 2}School of Electronic Science and Engineering\\

    Nanjing University\\
    Nanjing, 210023, China\\
    mengjunyi@smail.nju.edu.cn, frshen@nju.edu.cn
}

\begin{document}

\maketitle

\begin{abstract}
Federated LoRA fine-tuning enables parameter-efficient adaptation of pre-trained models without sharing private data, but suffers from two fundamental mismatches under heterogeneous client data: a structural aggregation mismatch caused by independently averaging LoRA factors, and a statistical collaboration mismatch caused by enforcing a single global adapter across divergent clients. 
To address these issues, we propose CF-LoRA, a clustered federated LoRA fine-tuning framework that combines decoupled factor aggregation with adaptation-aware client clustering. 
CF-LoRA first learns a globally shared $A$ factor while retaining personalized $B_i$ factors, then identifies clients with similar adaptation patterns based on the cosine similarity of their learned $B_i$ factors, and finally performs intra-cluster $B$-factor aggregation with a frozen $A$ factor. 
By decoupling LoRA factor aggregation, CF-LoRA preserves the low-rank structure and mitigates the structural aggregation mismatch, while adaptation-aware clustering promotes collaboration among clients with similar adaptation patterns and reduces negative transfer caused by statistical heterogeneity. 
Experiments on four language tasks and four vision datasets with RoBERTa and ViT show that CF-LoRA achieves the highest average accuracy in both modalities while communicating only one LoRA factor per optimization round.
\end{abstract}


\section{Introduction}

Large pre-trained models~\cite{touvron2023llama} have become a foundation for a broad range of language and vision applications, yet deploying them in specialized domains requires adaptation to task-specific data. 
Such data are often distributed across institutions or edge devices and cannot be centralized because of privacy, regulatory, and ownership constraints. 
Federated learning (FL)~\cite{li2020federated} enables collaborative adaptation without sharing raw data, but full-model fine-tuning incurs prohibitive computation and communication costs. 
Parameter-efficient fine-tuning (PEFT)~\cite{houlsby2019parameter}, particularly Low-Rank Adaptation (LoRA)~\cite{hu2022lora}, offers a natural remedy by freezing the pre-trained backbone and representing each weight update as the product of two trainable low-rank matrices, $\Delta W=BA$. 
Combining FL with LoRA therefore promises privacy-preserving and communication-efficient adaptation of large pre-trained models. 
However, effective federated LoRA fine-tuning requires answering two coupled questions: \emph{how should low-rank adaptations be aggregated, and with whom should each client aggregate?}

\begin{figure}[t]
	\centering
	\includegraphics[width=\linewidth]{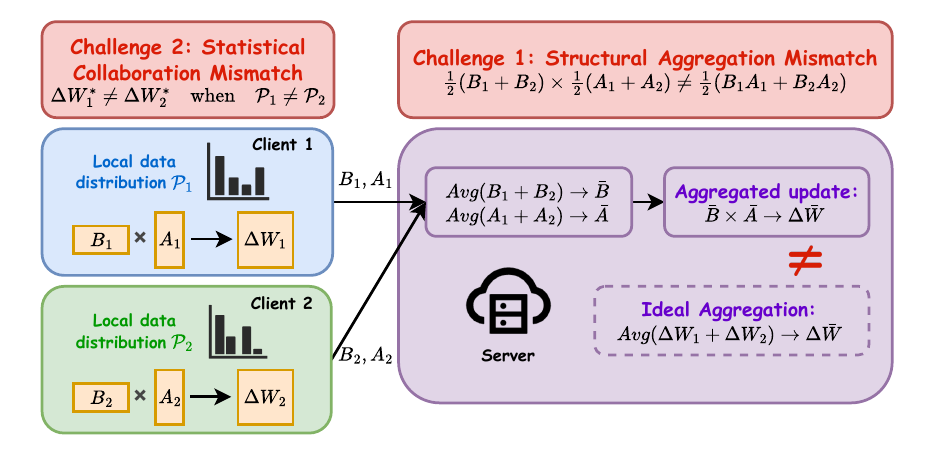}
    \caption{Two mismatches in federated LoRA fine-tuning.
    \textbf{Right}: the \emph{structural aggregation mismatch} arises because independently averaging the LoRA factors introduces spurious cross-client products, making the resulting update inconsistent with the ideal aggregated update.
    \textbf{Left}: the \emph{statistical collaboration mismatch} arises because enforcing global collaboration among clients with heterogeneous data distributions can lead to negative transfer.
}
    \label{fig:challenges}
\end{figure}

As illustrated in Figure~\ref{fig:challenges}, these questions arise from two mismatches. 
The first is a \emph{structural aggregation mismatch}. Let client $i$ learn LoRA factors $(A_i,B_i)$, and let $p_i$ denote its aggregation weight, with $\sum_i p_i=1$. Directly applying factor-wise federated averaging produces
\begin{equation}
\left(\sum_{i=1}^{m}p_iB_i\right)
\left(\sum_{i=1}^{m}p_iA_i\right)
\neq
\sum_{i=1}^{m}p_iB_iA_i,
\label{equ:error}
\end{equation}
because multiplying the averaged factors introduces spurious cross-client products $B_iA_j$ for $i\neq j$, which are absent from the ideal averaged update.
Directly aggregating the products $B_iA_i$ avoids this discrepancy but destroys LoRA's compact low-rank parameterization. 
The second is a \emph{statistical collaboration mismatch}. 
Under non-IID data, clients with different local data distributions can favor substantially different adaptations:
\begin{equation}
\begin{aligned}
\Delta W_i^\star
&= \operatorname*{arg\,min}_{\Delta W} F_i(W_0+\Delta W),\\
\Delta W_i^\star
&\not\approx \Delta W_j^\star
\quad\text{when}\quad
\mathcal{P}_i\not\approx\mathcal{P}_j,
\end{aligned}
\end{equation}
where $F_i$ and $\mathcal{P}_i$ denote the local objective and data distribution of client $i$, respectively.
Forcing all clients to share a single LoRA adaptation can therefore cause negative transfer and slow convergence~\cite{li2022federated}. 
Conversely, keeping every adaptation fully local avoids cross-client interference but discards transferable knowledge among clients with related distributions~\cite{sattler2020clustered}. 
Federated LoRA must thus determine not only \emph{how} to aggregate its factorized updates, but also \emph{with whom} each client should collaborate.

Existing federated LoRA methods typically address only one side of this coupled challenge. 
FFA-LoRA~\cite{sunimproving} and FedEx-LoRA~\cite{singhal2025fedex} mitigate the structural aggregation mismatch by freezing one LoRA factor or introducing a residual correction, respectively, but both retain a global collaboration pattern that overlooks client heterogeneity. 
FedSA-LoRA~\cite{guo2025selective} globally aggregates $A$ while keeping each $B_i$ fully personalized, reducing cross-client interference but missing beneficial collaboration among clients with similar adaptations. 
Meanwhile, clustered FL methods such as PACFL~\cite{vahidian2023efficient} form collaboration groups based on local data subspaces, but their grouping is detached from the adaptations learned by LoRA and does not resolve the structural mismatch of standard factor-wise aggregation. 
Consequently, most existing approaches provide either LoRA-compatible aggregation without heterogeneity-aware collaboration or selective collaboration without aggregation-consistent LoRA updates.

To bridge this gap, we exploit an asymmetric property of the two LoRA factors. 
Our motivating study reveals that the learned $A$ matrices remain highly consistent across clients, whereas the $B$ matrices become increasingly divergent as data heterogeneity grows. 
This observation, consistent with prior analyses of LoRA asymmetry~\cite{zhu2024asymmetry}, suggests that $A$ is well suited to capture globally shared information, while $B$ provides an adaptation-aware representation of client-specific characteristics. 
Based on this insight, we propose \textbf{CF-LoRA}, a clustered federated LoRA fine-tuning framework that jointly addresses the two questions above. 
CF-LoRA first learns a globally shared $A$ by aggregating only the $A$ factors while retaining locally personalized $B_i$ factors. 
It then clusters clients according to the pairwise cosine similarities between their learned $B_i$ factors.
Finally, it freezes the shared $A$ and aggregates $B$ only among clients within the same cluster. 
By aggregating only one factor in each optimization stage, CF-LoRA mitigates the mismatch caused by independently averaging client-specific $A$ and $B$ factors.
By restricting collaboration to adaptation-similar clients, CF-LoRA preserves transferable knowledge while reducing interference from heterogeneous clients.
In short, CF-LoRA answers \emph{how to aggregate} through decoupled factor aggregation and \emph{with whom to aggregate} through $B$-based client clustering.

Our main contributions are summarized as follows:
\begin{itemize}

\item \textbf{To address the structural aggregation mismatch, we propose a decoupled factor aggregation scheme.}
The scheme aggregates only one low-rank factor in each optimization stage, thereby mitigating the aggregation error caused by jointly averaging both factors while preserving LoRA's low-rank structure.

\item \textbf{To address the statistical collaboration mismatch, we introduce $B$-based adaptation-aware client clustering.}
The learned $B$ matrices are used to identify clients with similar adaptation patterns, enabling beneficial intra-group knowledge sharing while reducing interference from dissimilar clients.

\item \textbf{We develop CF-LoRA, a unified three-stage framework that couples decoupled factor aggregation with adaptation-aware client collaboration.}
CF-LoRA combines global $A$ learning with local $B_i$ personalization, $B$-based client clustering, and cluster-wise $B$ aggregation under a frozen shared $A$, thereby jointly addressing the structural aggregation mismatch and the statistical collaboration mismatch.

\item \textbf{Extensive experiments demonstrate the effectiveness and generality of CF-LoRA.}
CF-LoRA consistently outperforms strong federated LoRA baselines across language and vision benchmarks while communicating only one LoRA factor per optimization round.
\end{itemize}

\section{Related Work}
\subsection{Federated Learning under Data Heterogeneity}
Federated learning enables clients to train a shared model without exchanging raw data, but the non-IID distributions common in practice can slow convergence and impair the generalization of FedAvg. 
Optimization-oriented methods mitigate this problem by regularizing local training, as in FedProx~\cite{li2020fedprox}, or correcting client drift, as in SCAFFOLD~\cite{karimireddy2020scaffold}; nevertheless, they retain a single global model. 
Personalized federated learning~\cite{tan2022towards} instead learns client-specific models, but fully local personalization can forgo useful transfer among partially related clients. 
Clustered FL~\cite{sattler2020clustered} offers an intermediate collaboration granularity by training one model for each group of similar clients. 
Representative approaches infer groups through iterative model assignment~\cite{ghosh2022efficient} or client data-subspace similarity~\cite{vahidian2023efficient}. 
However, classical clustered methods maintain and communicate multiple full models, making their direct application to large pre-trained models expensive. 
This motivates combining heterogeneity-aware collaboration with parameter-efficient adaptation.

\subsection{PEFT and Federated LoRA Fine-Tuning}
Parameter-efficient fine-tuning (PEFT)~\cite{fu2023effectiveness} adapts pre-trained models while updating only a small parameter subset. 
Federated PEFT has been explored through lightweight adapters~\cite{chen2024feddat}, continuous prompts~\cite{guo2023promptfl}, and sparsely activated modules~\cite{wu2024fedfmsl}. 
Among these techniques, Low-Rank Adaptation (LoRA)~\cite{hu2022lora} is particularly attractive for federated fine-tuning because it freezes the pre-trained backbone and parameterizes weight updates using two trainable low-rank factors, thereby reducing both trainable parameters and communication overhead.

Existing federated LoRA methods have primarily explored aggregation consistency or heterogeneity-aware collaboration.
For aggregation consistency, FFA-LoRA~\cite{sunimproving} freezes one LoRA factor, FedEx-LoRA~\cite{singhal2025fedex} introduces a residual correction, FedSA-LoRA~\cite{guo2025selective} globally aggregates $A$ while retaining personalized $B$ factors, and FedRot-LoRA~\cite{zhang2026fedrotlora} aligns client factors through orthogonal transformations.
For heterogeneity-aware collaboration, FedLEASE~\cite{wang2025adaptive} constructs cluster-level LoRA experts with adaptive routing, while FedALT~\cite{bian2026fedalt} adaptively combines client-specific and shared LoRA adaptations.
In contrast, CF-LoRA couples adaptation-aware client clustering with decoupled factor aggregation, jointly addressing client heterogeneity and the structural aggregation mismatch.

\section{Method}

\subsection{Preliminaries}
We consider a federated learning setting with $m$ clients, where each client $i$ holds a local dataset $\mathcal{D}_i$ that cannot be shared due to privacy or regulatory constraints.
The data distributions across clients are potentially heterogeneous (non-IID), i.e., $\mathcal{D}_i \sim \mathcal{P}_i$ with $\mathcal{P}_i \neq \mathcal{P}_j$ for $i \neq j$.

Let $f(\cdot; \theta)$ denote a large pre-trained model with parameters $\theta$.
Instead of fine-tuning all model parameters, we adopt LoRA for parameter-efficient fine-tuning.
Specifically, for a weight matrix $W_0 \in \mathbb{R}^{d_{\mathrm{out}} \times d_{\mathrm{in}}}$ in the pre-trained model, LoRA introduces a low-rank update
\begin{equation}
\Delta W = B A,
\end{equation}
where $A \in \mathbb{R}^{r \times d_{\mathrm{in}}}$ and $B \in \mathbb{R}^{d_{\mathrm{out}} \times r}$ with rank $r \ll \min(d_{\mathrm{out}},d_{\mathrm{in}})$.
During training, the backbone parameters $\theta$ are frozen, and only the LoRA parameters $(A,B)$ are updated.

\begin{figure}[t]
	\centering
	\includegraphics[width=0.91\linewidth]{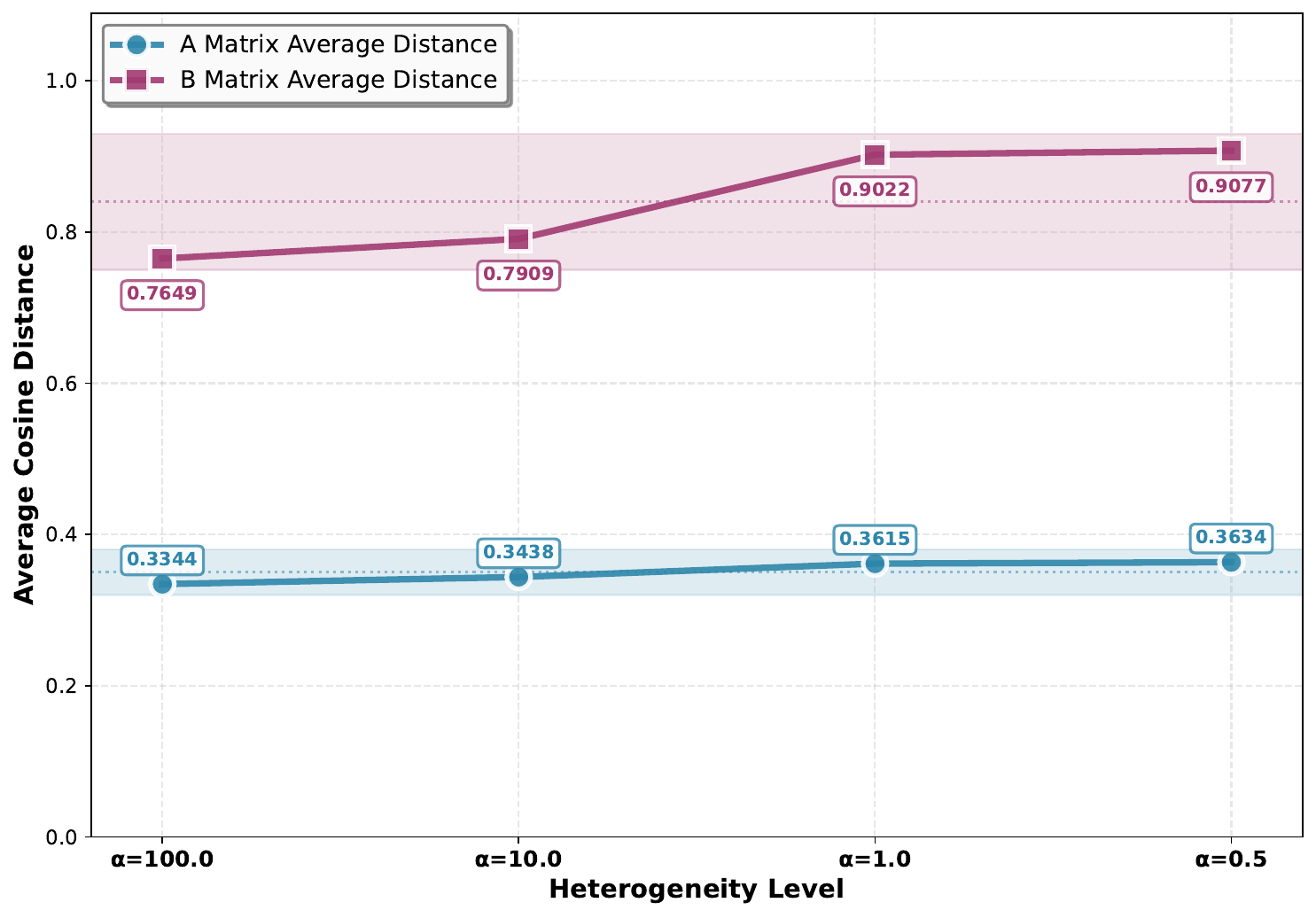}
    \caption{Cross-client cosine distances of learned LoRA factors on SST-2. Points show the mean pairwise cosine distance between flattened $A$ or $B$ factors under Dirichlet data partitioning; smaller $\alpha$ indicates greater heterogeneity. The $A$ factors remain comparatively similar across clients, whereas the $B$ factors diverge as heterogeneity increases.}
    \label{fig:cosine}
\end{figure}

\subsection{Motivating Observation}

Before introducing our method, we present a motivating empirical observation that highlights the different roles played by the two LoRA matrices, $A$ and $B$, in federated fine-tuning.

We conduct a toy FL study on the SST-2 task using RoBERTa-base as the backbone.
LoRA is applied to the attention layers, while the backbone parameters remain frozen.
All clients are initialized with the same LoRA parameters.
To simulate data heterogeneity, the training data are partitioned across 12 clients according to a Dirichlet distribution with varying concentration parameter $\alpha$, following standard non-IID settings in federated learning, where smaller $\alpha$ indicates higher heterogeneity.
Each client performs four local training epochs to update its LoRA parameters.

After training, we extract the learned LoRA matrices $(A_i, B_i)$ from each client.
To quantify cross-client consistency, we compute the pairwise cosine distance between flattened $A$ matrices and between flattened $B$ matrices, and report the averaged distances under different heterogeneity levels.
As shown in Figure~\ref{fig:cosine}, two consistent trends emerge.
First, the $A$ matrices exhibit significantly smaller cosine distances across clients than the $B$ matrices.
Second, as data heterogeneity increases, the cosine distance among $B$ matrices grows rapidly, whereas the distance among $A$ matrices remains relatively stable. 
These results suggest that the $A$ matrices primarily capture general information across clients, while the $B$ matrices focus on encoding client-specific adaptations that are highly sensitive to local data distributions. This observation is consistent with prior work analyzing the asymmetry of LoRA~\cite{zhu2024asymmetry}.
This empirical observation directly motivates our design of decoupling the aggregation of $A$ and $B$, and using the learned $B$ matrices as client representations for similarity-based clustering.
Additional analyses of the learned LoRA factors are provided in the supplementary material.

\begin{figure*}[htb]
	\centering
	\includegraphics[width=\linewidth]{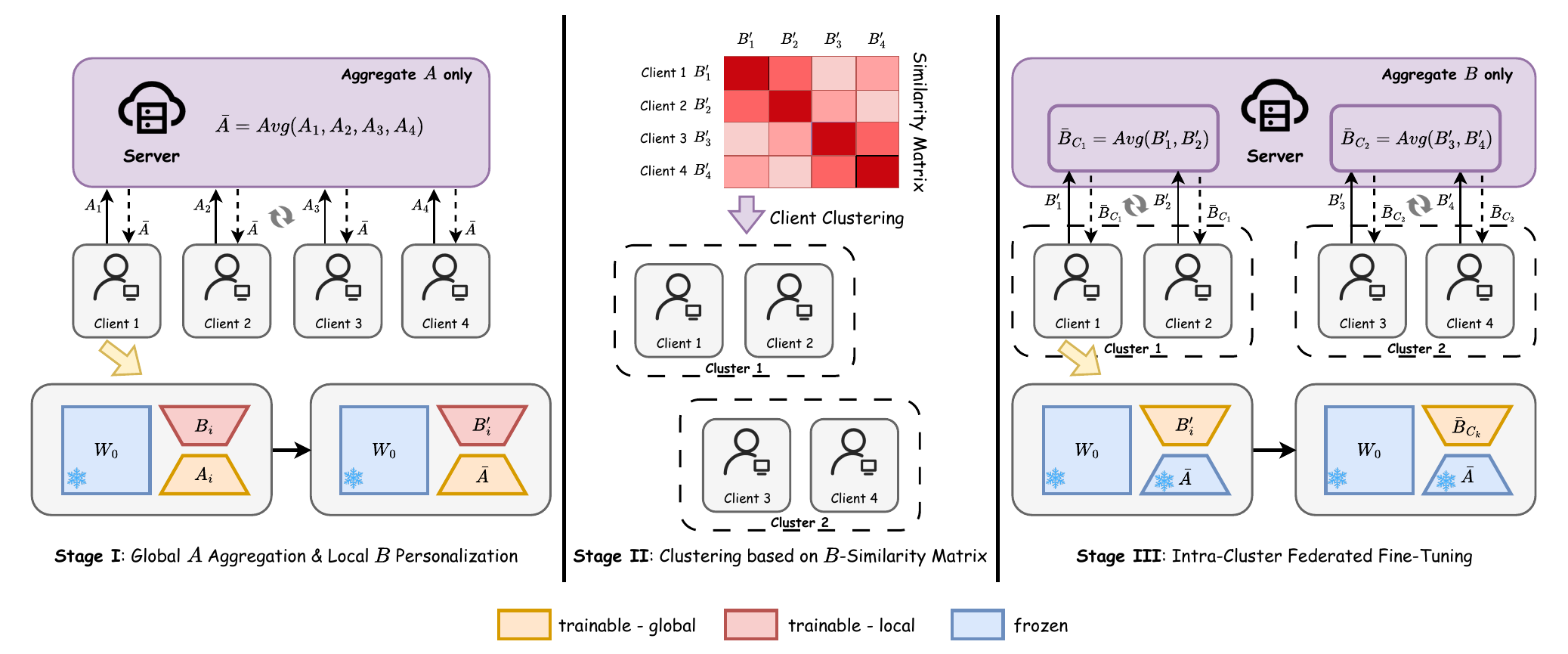}
    \caption{Overview of the CF-LoRA framework. CF-LoRA consists of three stages: (I) global aggregation of the LoRA $A$ matrices with local personalization of $B$, (II) client clustering based on the similarity of the learned $B$ matrices, and (III) intra-cluster federated fine-tuning by aggregating $B$ matrices with a frozen shared $A$. By aggregating only one low-rank matrix in each optimization stage, CF-LoRA avoids forming the product of independently averaged LoRA factors, while clustering similar clients effectively mitigates the negative effects of data heterogeneity.}
    \label{fig:framework}
\end{figure*}

\subsection{Overview of CF-LoRA}
Based on the above insights, we propose \textbf{CF-LoRA}, a clustered federated LoRA fine-tuning framework.
An overview of the proposed framework is illustrated in Figure~\ref{fig:framework}.

\subsubsection{Stage I: Global A Aggregation $\&$ Local B Personalization.} All clients start from the same initialization of LoRA parameters. In Stage I, each client locally optimizes its LoRA parameters $(A_i, B_i)$ while keeping the backbone model frozen. 

At communication round $t$, client $i$ performs $E$ steps of local optimization on its local dataset $\mathcal{D}_i$ to update its LoRA parameters:
\begin{equation}
(A_i^{(t)}, B_i^{(t)}) \leftarrow \text{LocalUpdate}\big((A_i^{(t-1)}, B_i^{(t-1)}); \theta, \mathcal{D}_i\big).
\end{equation}

After local training, only the $A$ matrices are uploaded to the server.
The server performs global aggregation by averaging:
\begin{equation}
\bar{A}^{(t)} = \frac{1}{m} \sum_{i=1}^{m} A_i^{(t)}.
\end{equation}

The aggregated matrix $\bar{A}^{(t)}$ is then broadcast to all clients and replaces their local $A_i^{(t)}$ for the next communication round, while the $B_i^{(t)}$ matrices remain entirely local.

This procedure is repeated for $T_1$ communication rounds, resulting in a globally shared low-rank matrix $\bar{A}$ that captures cross-client knowledge.
We denote the resulting local matrices at client $i$ after Stage~I as $B_i'$, which serve as compact client representations and are used for similarity-based clustering in the next stage.

In Stage~I, the server aggregates only the $A$ matrices, while each $B_i$ remains local to client $i$.
The resulting update at client $i$ is
\begin{equation}
W_0 + B_i \bar{A}
= W_0 + B_i \frac{1}{m}\sum_{j=1}^{m} A_j.
\end{equation}
Thus, Stage~I adopts a shared-$A$, personalized-$B_i$ parameterization and avoids multiplying two independently averaged LoRA factors, which causes the structural aggregation mismatch in Eq.~(\ref{equ:error}).

\subsubsection{Stage II: Clustering based on B-Similarity Matrix.} After Stage I, each client has learned a personalized $B_i'$ matrix that reflects its local data characteristics.
Since the $B_i'$ matrices are optimized on local datasets $\mathcal{D}_i$ under a shared backbone and a periodically synchronized $A$ matrix, they implicitly encode client-specific adaptations induced by the underlying data distributions $\mathcal{P}_i$.
Each client uploads its final Stage~I matrix $B_i'$ once to the server, which flattens these client representations and computes pairwise cosine similarities:
\begin{equation}
s_{ij} = \frac{\langle \mathrm{vec}(B_i'), \mathrm{vec}(B_j') \rangle}
{\|\mathrm{vec}(B_i')\| \, \|\mathrm{vec}(B_j')\|}.
\end{equation}

Based on the similarity matrix $\{s_{ij}\}$, we apply a clustering algorithm (e.g., hierarchical clustering) to partition the clients into $K$ clusters:
\begin{equation}
\{\mathcal{C}_1, \mathcal{C}_2, \ldots, \mathcal{C}_K\}.
\end{equation}

Clients within the same cluster are expected to share similar data distributions and thus benefit from collaborative fine-tuning, which helps mitigate the negative effects of data heterogeneity in subsequent federated fine-tuning.

\subsubsection{Stage III: Intra-Cluster Federated Fine-Tuning.} In Stage III, we perform federated fine-tuning independently and in parallel within each cluster.
For a given cluster $\mathcal{C}_k$, the globally aggregated matrix $\bar{A}$ obtained from Stage~I is frozen and shared by all clients in the cluster.

Before intra-cluster training begins, the server initializes a shared $B$ matrix for each cluster by averaging the final Stage~I matrices of its members:
\begin{equation}
\bar{B}_{\mathcal{C}_k}^{(0)}
= \frac{1}{|\mathcal{C}_k|}\sum_{i\in\mathcal{C}_k}B_i'.
\end{equation}
The server broadcasts $\bar{B}_{\mathcal{C}_k}^{(0)}$ to all clients in $\mathcal{C}_k$, which set $B_i'^{(0)}\leftarrow\bar{B}_{\mathcal{C}_k}^{(0)}$.
Thus, all clients within a cluster start Stage~III from the same cluster-level initialization.

At communication round $t=1,\ldots,T_2$, each client $i \in \mathcal{C}_k$ performs $E$ steps of local optimization on its local dataset $\mathcal{D}_i$ to update its $B$ matrix:
\begin{equation}
B_i'^{(t)} \leftarrow \text{LocalUpdate}\big(B_i'^{(t-1)};\, \bar{A}, \theta, \mathcal{D}_i\big).
\end{equation}

After local training, clients upload their $B_i'^{(t)}$ to the server, which aggregates them as:
\begin{equation}
\bar{B}_{\mathcal{C}_k}^{(t)} = \frac{1}{|\mathcal{C}_k|} \sum_{i \in \mathcal{C}_k} B_i'^{(t)}.
\end{equation}

The aggregated $\bar{B}_{\mathcal{C}_k}^{(t)}$ is then broadcast back to the clients in the same cluster and replaces their local $B$ matrix.
The intra-cluster federated optimization runs for $T_2$ rounds.

In Stage~III, the shared matrix $\bar{A}$ is frozen, and only the $B$ matrices are aggregated within each cluster.
As an illustrative example, consider a cluster with two clients.
The aggregated update is given by
\begin{equation}
W_0 + \bar{B}_{\mathcal{C}}\bar{A} = W_0 + \frac{1}{2}(B_1' + B_2') \bar{A}
= W_0 + \frac{1}{2}(B_1' \bar{A} + B_2' \bar{A}),
\end{equation}
which is again a linear combination of client updates under a fixed $\bar{A}$.
Because every client uses the same frozen $\bar A$, this expression is exactly the average of the clients' induced updates.
Therefore, no factor-wise aggregation mismatch arises in the final intra-cluster optimization.
The complete algorithm and convergence analysis are provided in the supplementary material.

\section{Experiments}
\subsection{Experimental Setup}
\subsubsection{Datasets and Data Partitioning.}
We evaluate the proposed method on different natural language understanding (NLU) benchmarks from the GLUE suite~\cite{wang2018glue}: MNLI-matched, MNLI-mismatched, MRPC, QQP, and SST-2.
These datasets cover a diverse set of tasks, including natural language inference, paraphrase identification, and sentiment classification, enabling a comprehensive evaluation under different task characteristics. 

To simulate data heterogeneity in federated learning, we partition the official training and validation sets across clients using the same class-wise Dirichlet proportions with $\alpha=1$, producing non-IID local training and test sets.
Unless otherwise specified, all experiments are conducted under this data partitioning scheme. 

\subsubsection{Baseline Methods.}
We compare CF-LoRA with a wide range of baselines covering different design choices for federated fine-tuning of large pre-trained models:
\begin{itemize}
\setlength{\itemsep}{0pt}

\item \textbf{FL-LoRA}: A straightforward combination of LoRA with the FedAvg algorithm, where both LoRA matrices $(A,B)$ are locally trained and independently averaged on the server (e.g., FedIT~\cite{zhang2024towards}).

\item \textbf{FFA-LoRA}~\cite{sunimproving}: A federated LoRA approach that freezes the $A$ matrix and only updates and aggregates the $B$ matrix, aiming to mitigate aggregation mismatch.

\item \textbf{FedEx-LoRA}~\cite{singhal2025fedex}: An extension of FL-LoRA that introduces an additional residual term $\Delta W_{res}$ on the server to correct the aggregation error caused by independently averaging $A$ and $B$ matrices.

\item \textbf{FedSA-LoRA}~\cite{guo2025selective}: A selective aggregation method that globally shares and averages only $A$ while keeping each client's $B$ locally personalized.

\item \textbf{FedRot-LoRA}~\cite{zhang2026fedrotlora}: Aligns client LoRA factors through orthogonal transformations before aggregation to mitigate rotational misalignment while preserving their induced updates.

\item \textbf{PACFL + FedIT}: A clustered federated LoRA baseline that applies PACFL~\cite{vahidian2023efficient} to cluster clients by computing principal angles between their local data subspaces, followed by FL-LoRA-based fine-tuning within each cluster.

\item \textbf{CF-LoRA}: The proposed method, which decouples the aggregation of $A$ and $B$ and performs similarity-based clustering using the learned $B$ matrices.

\end{itemize}

\begin{table*}[t]
\centering
\small
\setlength{\tabcolsep}{6pt}
\begin{tabular}{l *{6}{p{1.5cm}<{\centering}}}
\toprule
\textbf{Method} & \textbf{MNLI-m} & \textbf{MNLI-mm} & \textbf{MRPC} & \textbf{QQP} & \textbf{SST-2} & \textbf{Average} \\
\midrule
FL-LoRA (\scriptsize ICASSP'24)    & 86.83±0.06 & 86.56±0.08 & 87.25±0.25 & 89.61±0.05 & 94.76±0.07 & 89.00 \\
FFA-LoRA (\scriptsize ICLR'24)     & 85.08±0.03 & 85.24±0.08 & 85.62±0.14 & 88.06±0.01 & 94.11±0.18 & 87.62 \\
FedEx-LoRA (\scriptsize ACL'25)    & 86.67±0.14 & 86.53±0.03 & 86.93±0.37 & 89.50±0.04 & 94.57±0.26 & 88.84 \\
FedSA-LoRA (\scriptsize ICLR'25)   & 87.16±0.05 & 86.65±0.10 & 85.13±0.38 & 90.60±0.10 & 94.53±0.07 & 88.81 \\
FedRot-LoRA (\scriptsize ICML'26)  & 86.99±0.25 & 86.75±0.13 & 87.91±0.75 & 89.67±0.10 & 94.91±0.29 & 89.25 \\
PACFL + FedIT (\scriptsize AAAI'23)   & 86.60±0.11 & 86.26±0.03 & 87.27±0.25 & 90.73±0.03 & 95.04±0.13 & 89.18 \\
\rowcolor{gray!20} \textbf{CF-LoRA (Ours)} & \textbf{87.72±0.02} & \textbf{87.11±0.34} & \textbf{91.17±0.88} & \textbf{91.35±0.06} & \textbf{95.20±0.19} & \textbf{90.51} \\
\bottomrule
\end{tabular}
\caption{Performance of different methods on five GLUE evaluation sets under Dirichlet-based data partitioning with $\alpha=1$. MNLI-m and MNLI-mm denote the matched and mismatched evaluation sets of MNLI, respectively. We report accuracy ($\%$) where higher values indicate better performance. For all evaluation sets, we report accuracy evaluated across 3 runs with mean and standard deviation.
}
\label{tab:main_results}
\end{table*}

\subsubsection{Implementation Details.}
All methods are implemented using the RoBERTa-base backbone (125M parameters) from the HuggingFace Transformers library~\cite{wolf2020transformers}.
We simulate a federated learning environment with $12$ clients.
Each federated optimization round consists of $3$ local training epochs.
All baseline methods are trained for $50$ federated optimization rounds.
For CF-LoRA, we allocate $4$ rounds to the initial global training stage and $46$ rounds to the clustered federated training stage, resulting in the same total of $50$ federated optimization rounds.
As shown in the supplementary material, the client clustering assignments changes negligibly when the number of Stage~I rounds increases from 4 to 10, motivating the choice of $T_1=4$.
We apply LoRA to the query ($Q$) and value ($V$) projection matrices of the attention layers.
By default, LoRA uses rank $r=4$, scaling factor $8$, and dropout $0.1$.
We report sensitivity analyses for the LoRA rank $r$ in the supplementary material.

For initialization, the $A$ matrices are initialized using Kaiming initialization, while the $B$ matrices are initialized to zero.
All clients share the same initial LoRA parameters.
For CF-LoRA, client clustering is performed after Stage~I based on the learned LoRA $B$ matrices.
Specifically, each $B$ matrix is flattened and used as the client representation.
We compute pairwise cosine similarity between client representations and apply hierarchical clustering to group clients.
For PACFL + FedIT, clustering follows the original PACFL setting based on data subspace similarity.
For clustering-based methods, the number of clusters is set to $3$ for MNLI-m and MNLI-mm, $2$ for MRPC, $4$ for QQP, and $2$ for SST-2.
We report sensitivity analyses for the number of clusters $K$ in the supplementary material.
The FedRot-LoRA soft-rotation strength is set to $\lambda=0.4$.
All experiments are implemented in PyTorch and conducted on NVIDIA Tesla V100 GPUs.
All models are trained using the AdamW optimizer with a learning rate of $0.001$ and a warmup ratio of $0.05$.
We use cross-entropy loss for all tasks.
Each client is evaluated using its corresponding trained model; if a method learns only a single shared model, the shared model is used for all clients.
We report Top-1 accuracy as a test-set-size-weighted average across clients:
\begin{equation}
\mathrm{Acc}
=
\sum_{i=1}^{m}
\frac{n_i^{\mathrm{test}}}
{\sum_{j=1}^{m} n_j^{\mathrm{test}}}
\mathrm{Acc}_i,
\end{equation}
where $n_i^{\mathrm{test}}$ and $\mathrm{Acc}_i$ denote the number of test samples and the accuracy of client $i$, respectively.

\begin{table}[t]
\centering
\small
\begin{tabular}{lcc}
\toprule
\textbf{Method} & \textbf{Upload} & \textbf{Download} \\
\midrule
FL-LoRA        & $A + B$ & $A + B$ \\
FFA-LoRA       & $B$     & $B$ \\
FedEx-LoRA     & $A + B$ & $A + B + \Delta W_{res}$ \\
FedSA-LoRA     & $A$     & $A$ \\
FedRot-LoRA    & $A + B$ & $A + B$ \\
PACFL + FedIT     & $A + B$ & $A + B$ \\
\rowcolor{gray!20} \textbf{CF-LoRA (Ours)}  & $A$ \textbf{or} $B$ & $A$ \textbf{or} $B$ \\
\bottomrule
\end{tabular}
\caption{Comparison of uplink and downlink communication payloads across federated LoRA methods in each optimization round.}
\label{tab:comm_cost}
\end{table}

\subsection{Overall Performance}
\subsubsection{Better NLU Performance.}

Table~\ref{tab:main_results} reports the overall performance on five GLUE evaluation sets.
CF-LoRA achieves the highest accuracy on all evaluation sets and improves the average accuracy of FL-LoRA from 89.00\% to 90.51\%.
Compared with FedRot-LoRA, the strongest recent baseline, CF-LoRA achieves higher accuracy on all five evaluation sets and improves the average accuracy by 1.26 percentage points.
These results demonstrate the benefit of jointly addressing aggregation mismatch and client heterogeneity.

Existing methods typically address only one aspect of the problem.
FFA-LoRA limits adaptation by freezing $A$; FedEx-LoRA and FedRot-LoRA improve aggregation consistency but retain global collaboration; FedSA-LoRA preserves personalized $B$ factors but overlooks collaboration among similar clients; and PACFL + FedIT clusters clients based on local data subspaces, making its grouping dependent on dataset-specific characteristics and leading to varying clustering effectiveness across datasets. 
For example, it improves over FL-LoRA on QQP and SST-2 but underperforms it on both MNLI evaluation sets.
In contrast, CF-LoRA combines decoupled factor aggregation with adaptation-aware clustering, leading to consistent improvements across tasks.

\subsubsection{Communication Efficiency.}
Table~\ref{tab:comm_cost} summarizes the parameters transmitted by different federated LoRA methods in a single optimization round.
FL-LoRA, PACFL + FedIT, and the recent FedRot-LoRA transmit both $A$ and $B$ in each direction, while FedEx-LoRA additionally downloads a residual update.
CF-LoRA uploads and downloads only $A$ during Stage~I and only $B$ during Stage~III, so it never transmits both factors in one optimization round.
Because $A$ and $B$ have equal size in our experiments, CF-LoRA communicates one factor-equivalent in each direction per optimization round, compared with two for methods that communicate $A+B$.
This reduces the iterative per-round LoRA-parameter payload by 50\% in both the uplink and downlink.
Although FFA-LoRA and FedSA-LoRA also communicate a single factor per round, Table~\ref{tab:main_results} shows that CF-LoRA attains substantially higher average accuracy while retaining the same per-round payload class.
CF-LoRA thus provides a favorable accuracy--communication trade-off while communicating only one LoRA factor per round.

\begin{figure}[t]
\centering
\includegraphics[width=\linewidth]{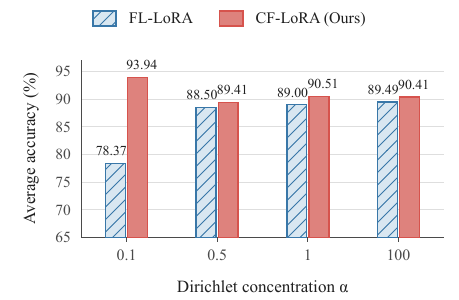}
\caption{Mean accuracy across the five GLUE evaluation sets under different Dirichlet heterogeneity levels. Smaller $\alpha$ indicates stronger heterogeneity, and values above the bars report the corresponding mean accuracies.}
\label{fig:heterogeneity}
\end{figure}

\subsection{Robustness Analysis}

\subsubsection{Effect of Data Heterogeneity.}
Figure~\ref{fig:heterogeneity} compares CF-LoRA and FL-LoRA under different levels of Dirichlet heterogeneity.
CF-LoRA consistently outperforms FL-LoRA, with the largest margin observed at $\alpha=0.1$, where the average accuracies are 93.94\% and 78.37\%, respectively.
A likely explanation is that the evaluated GLUE tasks contain only two or three labels, so an extremely heterogeneous partition may leave some clients with samples from only one or a few classes.
In this case, CF-LoRA groups clients with similar adaptation patterns, allowing compatible updates to reinforce one another through intra-cluster aggregation.
In contrast, global aggregation mixes updates from clients with substantially different label distributions, resulting in stronger interference.
These results show that CF-LoRA remains effective across different non-IID data settings.
Complete per-dataset results are provided in the supplementary material.

\begin{table}[t]
\centering
\small
\setlength{\tabcolsep}{3pt}
\begin{tabular}{c|lcccccc}
\toprule
\textbf{Num} & \textbf{Method} & M-m & M-mm 
& MRPC & QQP & SST-2 & \textbf{Avg.} \\
\midrule
\multirow{2}{*}{12}
& FL-LoRA
& 86.83 & 86.56 & 87.25 & 89.61 & 94.76 & 89.00 \\
& \textbf{Ours}
& \textbf{87.72} & \textbf{87.11} & \textbf{91.17}
& \textbf{91.35} & \textbf{95.20} & \textbf{90.51} \\
\midrule
\multirow{2}{*}{50}
& FL-LoRA
& 86.55 & \textbf{86.58} & 79.17 & 89.58 & 94.38 & 87.25 \\
& \textbf{Ours}
& \textbf{86.58} & 86.14 & \textbf{88.93}
& \textbf{89.80} & \textbf{94.69} & \textbf{89.23} \\
\bottomrule
\end{tabular}
\caption{Accuracy (\%) with 12 and 50 clients on the five GLUE evaluation sets.}
\label{tab:client_scalability}
\end{table}

\subsubsection{Effect of Federation Size.}
Table~\ref{tab:client_scalability} evaluates CF-LoRA and FL-LoRA as the number of clients increases from 12 to 50.
While the average accuracy of both methods decreases, CF-LoRA exhibits a smaller drop than FL-LoRA (1.28 versus 1.75 percentage points) and maintains a clear performance advantage, demonstrating greater robustness to a larger federation.

\subsubsection{Effect of Client Grouping.}
Table~\ref{tab:clustering_ablation} evaluates different grouping strategies using the same learned LoRA $B$-based representations.
Our hierarchical clustering strategy achieves the highest average accuracy of 90.51\%, outperforming K-means, spectral clustering, and random grouping by 1.43, 1.96, and 2.52 percentage points, respectively.
The substantial gain over random grouping highlights the importance of similarity-aware collaboration, while the improvements over alternative clustering methods indicate that hierarchical clustering better captures the structure of the learned client representations. 
Moreover, the clustering module is decoupled from the overall framework and can be readily replaced by more advanced clustering methods.

\begin{table}[t]
\centering
\small
\setlength{\tabcolsep}{4pt}
\begin{tabular}{lcccccc}
\toprule
\textbf{Clustering} & M-m & M-mm 
& MRPC & QQP & SST-2 & \textbf{Avg.} \\
\midrule
Random  
& 86.24 & 86.46 & 84.06 & 89.13 & 94.05 & 87.99 \\

K-means
& 86.10 & 86.54 & 89.70 & 89.00 & 94.05 & 89.08 \\

Spectral 
& 85.91 & \textbf{87.58} & 86.27 & 88.69 & 94.28 & 88.55 \\

Hierarchical
& \textbf{87.72} & 87.11 & \textbf{91.17}
& \textbf{91.35} & \textbf{95.20} & \textbf{90.51} \\
\bottomrule
\end{tabular}
\caption{Ablation of client grouping strategies using the same LoRA $B$-based representations. We report accuracy (\%).}
\label{tab:clustering_ablation}
\end{table}

\subsection{Evaluation on Vision Tasks with ViT-B/16}\label{vision}

To further demonstrate the generality of CF-LoRA beyond language models, we evaluate our method on vision tasks using ViT-B/16~\cite{dosovitskiy2021an} as the pre-trained backbone.
Specifically, we conduct federated fine-tuning experiments on four image classification benchmarks: Flowers102~\cite{nilsback2008automated}, DTD~\cite{cimpoi2014describing}, UCF101~\cite{soomro2012ucf101}, and Caltech101~\cite{fei2004learning}.
Detailed experimental settings are provided in the supplementary material.

Table~\ref{tab:downstream_results} summarizes the performance of different federated LoRA methods on four vision datasets.
CF-LoRA achieves the best performance across all datasets, demonstrating its effectiveness on diverse visual downstream tasks.
Together with the language results, these findings confirm that decoupled factor aggregation and $B$-based clustered collaboration generalize across model architectures and modalities, from RoBERTa-based language understanding to ViT-based image classification.

\begin{table}[t]
\centering
\small
\setlength{\tabcolsep}{4pt}
\begin{tabular}{lccccc}
\toprule
\textbf{Method} & Flowers & DTD & UCF
& Caltech & \textbf{Average} \\
\midrule
FL-LoRA
& 97.53 & 61.76 & 74.43 & 93.96 & 81.92 \\

FFA-LoRA
& 92.97 & 51.83 & 58.82 & 80.12 & 70.94 \\

FedEx-LoRA
& 97.82 & 63.18 & 77.42 & 94.12 & 83.14 \\

FedSA-LoRA
& 93.79 & 66.19 & 79.18 & 90.47 & 82.41 \\

FedRot-LoRA
& 98.23 & 62.77 & 74.81 & 94.12 & 82.48 \\

PACFL + FedIT
& 97.50 & 63.48 & 77.29 & 93.65 & 82.98 \\

\rowcolor{gray!20}
\textbf{CF-LoRA (Ours)}
& \textbf{98.31} & \textbf{66.53} & \textbf{79.97}
& \textbf{94.25} & \textbf{84.77} \\
\bottomrule
\end{tabular}
\caption{Performance comparison on vision datasets using ViT-B/16 as the pre-trained backbone. We report classification accuracy (\%).}
\label{tab:downstream_results}
\end{table}

\subsection{Limitations}
CF-LoRA currently uses a predefined number of clusters $K$ and performs hierarchical clustering only once after Stage~I.
This static design avoids repeated representation extraction and re-clustering, thereby keeping the additional computational overhead low.
However, a prespecified $K$ may not always reflect the intrinsic grouping structure of the clients, while a one-time partition cannot adapt when client distributions evolve, new clients join, or new data become available.
Although we use hierarchical clustering in the current implementation, the LoRA $B$ matrix representation is not inherently tied to a fixed-$K$ grouping rule.
Adaptive clustering methods that do not require a pre-specified number of clusters could instead use the $B$-based client similarities to infer the grouping structure and update the client assignments when needed.
Evaluating such extensions, as well as their trade-off between adaptivity, stability, and additional computation, is left for future work.

\section{Conclusion}

In this paper, we proposed CF-LoRA for federated LoRA fine-tuning under heterogeneous client data.
CF-LoRA addresses the structural aggregation mismatch through decoupled factor aggregation and mitigates the statistical collaboration mismatch through $B$-based client clustering.
Experiments on language and vision tasks show that CF-LoRA achieves the best average performance, remains robust under different heterogeneity levels and federation sizes, and communicates only one LoRA factor per optimization round.
These results demonstrate the effectiveness and generality of CF-LoRA.

\bibliography{aaai2027}


\end{document}